\documentclass[11pt,a4paper]{article}
\usepackage[T4,T1]{fontenc}

\usepackage{tikzsymbols}
\usepackage{enumitem}
\usepackage{amsmath}

\newenvironment{tfour}{\fontencoding{T4}\selectfont}{}
\usepackage{newunicodechar}
\usepackage{latexsym}
\usepackage[hyperref]{eacl2021}
\usepackage{times}
\usepackage{latexsym}

\usepackage{url}
\usepackage{xcolor}
\usepackage{graphicx}
\usepackage{tabularx}
\usepackage{subcaption}
\usepackage{array}
\usepackage{multirow}
\usepackage{booktabs}
\usepackage{microtype}

\aclfinalcopy % Uncomment this line for the final submission
\newunicodechar{ɔ}{\fon}
\newunicodechar{ɖ}{\fon}

\title{AfriVEC: Word Embedding Models for African Languages. Case Study of Fon and Nobiin}

\author{Bonaventure F. P. Dossou\\
  Jacobs University Bremen\\
  \texttt{f.dossou@jacobs-university.de} \\\And
  Mohammed Sabry \\
  University of Khartoum \\
  \texttt{mhmd.sabry.ab@gmail.com} \\}

\date{}

\begin{document}
\maketitle
\begin{abstract}
From Word2Vec to GloVe, word embedding models have played key roles in the current state-of-the-art results achieved in Natural Language Processing. Designed to give significant and unique vectorized representations of words and entities, those models have proven to efficiently extract similarities and establish relationships reflecting semantic and contextual meaning among words and entities. African Languages, representing more than 31\% of the worldwide spoken languages, have recently been subject to lots of research. However, to the best of our knowledge, there are currently very few to none word embedding models for those languages words and entities, and none for the languages under study in this paper. After describing Glove, Word2Vec, and Poincaré embeddings functionalities, we build Word2Vec and Poincaré word embedding models for Fon and Nobiin, which show promising results. We test the applicability of transfer learning between these models as a landmark for African Languages to jointly involve in mitigating the scarcity of their resources, and attempt to provide linguistic and social interpretations of our results. Our main contribution is to arouse more interest in creating word embedding models proper to African Languages, ready for use, and that can significantly improve the performances of Natural Language Processing downstream tasks on them. The official repository and implementation is at: \url{https://github.com/bonaventuredossou/afrivec}
\end{abstract}
\section{Introduction}
Word Embedding models are very useful in Natural Language Processing downstream tasks and got modernized usage with learning paradigms like zero-shot learning \citep{zeroshot}, addressing labels representation problems in both image and text classification tasks \citep{norouzi2014zeroshot,esann16zeroshot}. This makes it for the sake of scientific research on African Languages very important to consider, because of their scarse data resources. Throughout this paper, our main contribution is to provide standardization and evaluation guidelines to any research on the space, through our methods and experiments.

Fon, and Nobiin are the two African Indigenous Languages (ALs) chosen for this study. Fon is a native language of Benin Republic, spoken in average by more than 2.2 million people in Benin, in Nigeria, and Togo. Nobiin is native to Northern Sudan and Southern Egypt, spoken in average by a million of people. Both languages cover a wide differential range of cultures as the speakers are from Western and Northern Africa.

Fon alphabet is based on the latin alphabet, with the addition of the letters so \begin{tfour}\m{o}\end{tfour}, \begin{tfour}\m{d}\end{tfour}, \begin{tfour}\m{e}\end{tfour}, and the digraphs gb, hw, kp, ny, and xw. There are 10 vowel phonemes in Fon: 6 said to be $closed$ [i, u, ĩ, ũ], and 4 said to be $opened$ [(\begin{tfour}\m{e}\end{tfour}, \begin{tfour}\m{o}\end{tfour}, a, ã]. There are 22 consonants (m, b, n, \begin{tfour}\m{d}\end{tfour}, p, t, d, c, j, k, g, kp, gb, f, v, s, z, x, h, xw, hw, w). \par Nobiin alphabet is primarily based on greek alphabet with some meroitic characters, but in most of the resources and modern usage of the language, it uses the following schema of 28 letters: there are 10 vowels: 5 are $opened$: (a, e, i, o, u) and 5 are $closed$: (â, ê, î, ô, û). There are 18 consonants: (b, d, f, g, h, j, k, m, n, r, s, t, w, y, sh, ch, gn, ģ).

Word Embedding (WE) modeling is an approach that provides a dense vector representation of words and captures something about their meaning. The goal of embedding methods is to organize symbolic objects (words, entities, concepts etc.) in a way such that their similarities in the embedding space reflects their semantic or functional similarities. WEs models are improvements of naive bag-of-word (BOW) modeling that relies on statistics like word counts and frequencies to create large sparse vectors, describing documents but not the meaning of the words.

WEs work by using an algorithm to train a set of fixed-length dense and continuous-valued vectors based on a large corpus of text. Each word is represented by a point in the embedding space and these points are learned and moved around based on the words that surround the target word. Their massive use in text representations is one of the key methods that led to breakthrough performances in many fields of natural language processing like machine translation, and named entity recognition, just to mention a few. The most famous WEs are GloVe \cite{glove}, Word2Vec \cite{word2vec}, and Poincaré Embeddings \cite{pointcare}. 

Additionally, we would also like to highlight BERT embeddings \citep{bert_embeddings} from Transformers, which are ubiquitous nowadays in Natural Language Processing, and have improved systems performances. However, transformer-based architectures require a lot of computing power and data, and as such they may not be suitable for small datasets (which is the case in the current paper) or to researchers that do not have access to GPUs, whereas Word2vec and Poincare and Glove are not computationally expensive.

\section{Word Embeddings: Related Works}
\label{related_work}
\subsection{GloVe}
GloVe, or Global Vectors for Word Representation, is an approach to capture the meaning of one word embedding towards the corpus (set of documents - a document is a sentence). The GloVe model trains on global co-occurrence counts of words and makes a sufficient use of statistics by minimizing least-squares error. This produces a word vector space with meaningful substructure, that sufficiently preserves words similarities with vector distance.
The probability that word at a given index $t'$ occurs in the context of a word $t$ is defined as: 
\begin{center}
$P_{tt'} =  \frac{A_{tt'}}{A_{t}}$ (1)
\end{center}
where $A$ is the words co-occurrence matrix. Each entry of $A$, is the number of times word $t'$ occurs in the context of word $t$.
The function F, encoding the information about the ratio co-occurrence between two vectors is defined as:
\begin{center}
$F(v_{t} - v_{t'}; \hat{v_{k}}) = \frac{P_{tk}}{P_{t'k}}$ (2)
\end{center}
where $v_{t}$, $v_{t'}$ are word vectors with indices $t$ and $t'$, and $\hat{v_{k}}$ a context vector with index $k$.
The left arguments of F are vectors and the right side is a scalar. Since F could be taken as a complicated parameterized function like a neural network for instance, doing so would obfuscate the linear structure we are trying to capture. To avoid this issue, we can first take the dot product of the arguments, preventing F from mixing the vector dimensions in undesirable ways. The equation (2) becomes then:
\begin{center}
$F((v_{t} - v_{t'})^{T}\hat{v_{k}}) = \frac{P_{tk}}{P_{t'k}}$ (3)
\end{center}
Since in the word-word co-occurrence matrix, the distinction between context words and standard words is arbitrary, probabilities ratio are replaced and the equation (3) becomes:
\begin{center}
$F((v_{t} - v_{t'})^{T}\hat{v_{k}}) = \frac{F(v_{t}^{T}\hat{v_{k}})}{F(v_{t'}^{T}\hat{v_{k}})}$ (4)
\end{center}
In the original paper, \citet{glove} set the equation (5) and solved it for $F$ being the \textit{exponential function}:
\begin{center}
$F(v_{t}^{T}\hat{v_{k}}) = P_{tk} = \frac{A_{tk}}{A_{t}}$ (5)
\end{center}
The final solution of the equation (5) is:
\begin{center}
$v_{t}^{T}\hat{v_{k}} + b_{t} + \hat{b_{k}} = log(A_{tk})$ (6)
\end{center}
where $b_{t}$ and $\hat{b_{k}}$ are respective bias for $v_{t}$ and $\hat{v_{k}}$, added to restore symmetry.

Finally the loss function to minimize is hence a linear regression function defined as:
\[ J = \sum_{t, t'=1}^{V} (f(A_{tt'}) v_{t}^{T}\hat{v_{t'}} + b_{i} + \hat{b_{j}} - log(A_{tt'}))^{2}, \]with $V$ being the size of the WE's vocabulary.
$f(A_{tt'})$ is a pre-defined weighting function, that should be continuous, non-decreasing and relatively small for large values of the argument. Obviously, there are infinite functions that could be constructed to satisfy these criterias.
The authors \citet{glove} used the function $f$ defined as:
\begin{center}
$f (x) =
\begin{cases}
        (\frac{x}{x_{max}})^{\alpha}, x < x_{max}\\
        1, x \geq x_{max}
    \end{cases}
$
\newline \newline where $\alpha \in (0;1)$
\end{center}
\subsection{Word2Vec}
There are 3 different types of Word2Vec parameter learning, and all of them are based on neural network models \cite{word2vec}.
\subsubsection{One-Word Context}
This approach is known as Continuous Bag-Of-Word (CBOW). The main idea is the consideration of a single word per context i.e. we have to predict one word given only one word. The input of the neural network in this context is a one-hot encoded vector of size (V,) followed by a hidden layer of size N with an input hidden layer weights matrix W of size V×N, and an output layer weights matrix W’ of size N×V, with softmax activation function. The objective here is to compute as probability, the vector representation of the word with index $i$:
\begin{center}
$p(w_{j}|w_{i})$
\end{center}
Let $a$ be our input vector filled with zeros, and a single 1 at the position t. The hidden vector $h$ is computed with the formula below:
\begin{center}
$h = W^{T}a = z^{T}_{w_{i}}$
\end{center}
where $z$ is the output vector of the word $w_{i}$. We can look at $h$ as the «input vector» of the word $a$. At the next step, we take the vector $h$ and apply a matrix multiplication similar to the previous one:
\begin{center}
$o_{j} = z'^{T}_{w_{j}}h$
\end{center}
where $z’$ is the output vector of the word $w_{j}$ with. This multiplication is performed for every entry $o$ with index $j$.
The $softmax$ activation is defined as followed:
\begin{center}
$p(w_{j}|w_{i}) = y_{j} = \frac{e^{o_{j}}}{\sum_{j'=1}^{V} e^{o_{j'}}}$
\end{center}
\subsubsection{Multi-Word Context}
The concept of Multi-word Context is very similar to the concept of CBOW. The only difference is that we want to capture the relationship between our target word and other words from the corpus. The probability distribution is then defined as:
\begin{center}
$p(w_{i}|w_{1,1},...,w_{1,c})$
\end{center}
obtained by changing the hidden layer function to:
\begin{center}
$h = \frac{1}{C}(\sum_{i=1}^{C} a_{i})$
\end{center}
The optimization function becomes then: 
\begin{center}
$-log(p(w_{i}|w_{1,1},...,w_{1,c}))$
\end{center}
\subsubsection{Skip-Gram Model}
The concept of Skip-Gram Model is opposite to the Multi-Word Model: the task is to predict \textit{$c$} context words having one target word on the input. The process of Skip-Gram is the reverse procedure of the Multi-Word Context \cite{word2vec}. The optimization function is then defined as followed:
\[ \frac{1}{T} \sum_{t=1}^{T} \sum_{-c\leq i \leq c} log(p(w_{t+i}|w_{t}))\]\textit{$-c$} and \textit{$c$} are limits of the context window and $w_{t}$ is a word at index $t$. \textit{$T$} is the total number number of words in the vocabulary.
The hidden vector $h$ is computed the same way as in the case of CBOW and Multi-Word Context. The output layer is computed with:
\begin{center}
$o_{c,j} = o_{j} = z'^{T}_{w_{j}}h$
\end{center}
and the activation function is defined as followed:
\begin{center}
$p(w_{c,j}=w_{j,c}|w_{i}) = y_{c,j} = \frac{e^{o_{c,j}}}{\sum_{j'=1}^{V} e^{o_{j'}}}$
\end{center}

\subsection{Poincaré Embeddings}
The concept of Poincaré Embeddings uses hyperbolic geometry to capture hierarchical complexities \citep{barabasi} and properties of the words that can not be captured directly in Euclidean space. There is a need to use such kind of geometry together with Poincaré ball to capture the fact that distance from the root of the tree to its leaves grows exponentially with every new child, and hyperbolic geometry is able to represent this property. Hyperbolic geometry studies non-Euclidean spaces of constant negative curvature. Its main 2 axioms and theorems are:
\setlist{nolistsep}
\begin{itemize}[noitemsep]
   \item $\forall$ line $a$ and $\forall$ point $p$ $\not\in$ a, there are at least two distinct parallels passing through $p$.
   \item all triangles have angles sum less than 180 degrees.
\end{itemize}

For 2-dimensional hyperbolic space, both the area $s$ and length $l$ of a circle, grow exponentially. The are defined with the following formulas:

\begin{center}
$l = 2\pi sinh(r)$ and $s = 2\pi (cosh(r)-1)$
\end{center}
where \textit{$r$} denotes the radius. The Poincaré Ball is defined then as:
\begin{center}
$B^{d} = \{x \in R^{d}| ||x|| < 1\}$
\end{center}
The distance measure between 2 WEs $t$ and $t'$ is defined as:
\begin{center}
$d(t, t') = arcosh(1 + 2\frac{||t-t'||^{2}}{(1- ||t||^{2})((1- ||t'||^{2})})$.
\end{center}
\citet{pointcare} argued that, this measure allows not only to capture effectively the similarity between the two WEs but also preserves their hierarchy (through their norm).
\paragraph{Relevant word embeddings works on African Languages:}
In this regard, very few explorations have been done, to the best of our knowledge. However it worths mentioning the work of \citet{adelani} which introduced massive and curative embeddings for Yoruba and Twi, two other low-resourced African Languages. The contextual word embeddings obtained have been used to evaluate multilingual BERT on a named entity recognition task.
Alternatively, \citet{isiZulu_wordEmbeddings} albeit on a very small monolingual English
web text corpus, explored the use of word embeddings in the synthesis of isiZulu-to-English code-switch bigrams used to augment sparse language model training data.
\section{Creation of our Word Embedding: Case Study of Fon and Nobiin Languages}
\label{word2vec_section}
\subsection{Contextualization}
A lookup at Google's Word2Vec model, can allow us to define queen as followed: $queen = (king - man) + woman$. $Woman$ and $Man$ are genders, while $queen$ and $king$ could be referred as titles. However, this equation, gives a meaningful and unique representation to the word $queen$, and its similarities with the words $king$, $man$ and $woman$.

In the context of ALs, to the best of our knowledge, there are no such models capable of capturing high-level relation between words and entities. \citet{adelani} also highlighted the difficulty of the evaluation on low-resourced languages, of several architectures that are capable to learn semantic representations from unannotated data, more suitable to high-resourced languages, that have a smorgasbord of tasks and test sets to evaluate on.

In general, this lack of research is coupled with the scarcity of the data and the morphological richness of ALs. Many studies including \cite{orife,ffr}, regarding the way to handle the textual data of African languages, have showed the importance of diacritics in the challenging task of creating effective and robust neural machine translation systems, and Natural Language Processing (NLP) tools for ALs. Therefore, during the preprocessing of the data used for this study, we made sure to keep the words diacritics, in order to not lose any meaningful information.

We aim to investigate the possibilities of creating proper word embedding models for a better and meaningful representations of ALs words and entities; models that could capture exclusive relation between words of the same language. Moreover, the results of our work, could be an effort to creating or improving Named Entity Recognition (NER) models for ALs. For each of the languages selected (Fon and Nobiin), we tried the following:

\begin{itemize}[noitemsep]
   \item create a word2vec embedding model, and test its capacity to establish relationships or similarities among words.
   
   %On the other hand, we want also to see how words relationships and similarities can reflect the culture. For instance, how is the word $mother$ related to home tasks, as in Africa most of women are cattle breeders and housekeeper. So the most similar to word $mother$ might be $cooking$, $cleaning$, or any other home tasks.%

   \item build a Poincaré WE model and use it to predict the types of entities.
\end{itemize}
To promote reproducibility, and further improvements as well as the use of our findings in NLP tasks on ALs. The datasets, and models source code will be open-sourced, and contributions are welcome. 
\subsection{Fon and Nobiin Word2Vec Embedding Models}
For investigating how well a Word2Vec model architecture could give provide vectorized representations of the words of our dataset, as well as establish the relationship between them, we created a basic Word2Vec embedding model using the library $gensim$ \footnote{https://radimrehurek.com/gensim/models/word2vec.html}. In our context we used the CBOW as training algorithm, which is a feed-forward Neural Network Language Model, where the non-linear hidden layer is removed and the projection layer is shared for all words. This ensures that the continuous distributed representations of words in their respective contexts are used.
\begin{figure}[t]
\includegraphics[width=\linewidth, height=70mm]{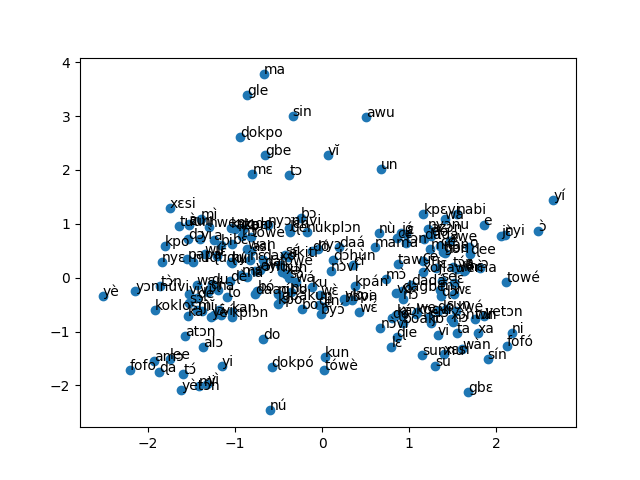}
\centering
\caption{\label{fig1} Visualization of Fon Word2Vec Embedding Space}
\end{figure}
\begin{figure}[t]
\includegraphics[width=\linewidth, height=70mm]{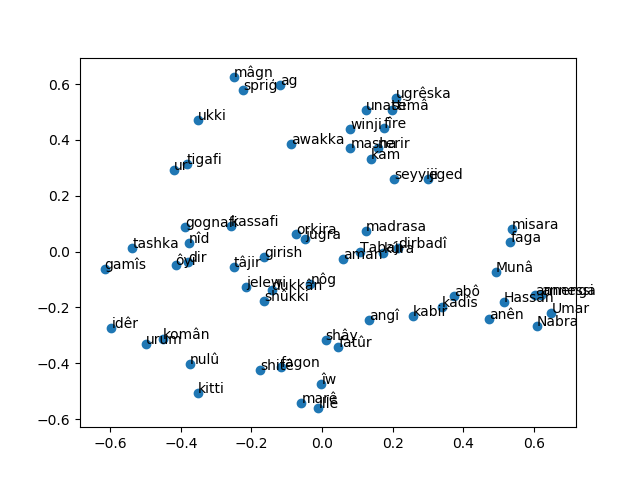}
\centering
\caption{\label{fig2}Visualization of Nobiin Word2Vec Embedding Space}
\end{figure}

%\begin{figure*}[t]
%\begin{minipage}[b]{1\linewidth}
%\centering
%\subfloat[]{\includegraphics[width=7cm]{family_word_embedding.png}}\qu%ad
%\subfloat[]{\includegraphics[width=7cm]{NobinWord2Vec.png}}
%\caption{\label{fig1}Visualization of (a) Fon and (b) Nobiin Word %Embeddings Space Word Vectors Using PCA}
%\end{minipage}
%\end{figure*}
\subsubsection{ Word2Vec for Fon}
For Fon, as a starting point, we chose to focus on the family domain (context) where words are dad, mum, sister, brother, son, daughter. From the FFR parallel dataset \citep{ffr, ffr_dataset}, we filtered and extracted, and manually cleaned Fon sentences containing the keywords mentioned above. The resultant dataset contains 739 sentences, with an average of 8 words per sentence. The following parameters have been used to create and train the Word2Vec model for Fon:
\begin{itemize}[noitemsep]
   \item size: the number of dimensions of the embedding, set to $100$.
   \item \text{min\char`_count}: the minimum count of words to consider when training the model, set to $5$.
   \item $\alpha$: the learning rate set to $0.5$.
   \item window: the maximum distance between a target word and words around the target word, set to $5$.
   \item workers: the number of threads to use while training, set to $3$.
   \item sg: the training algorithm; $0$ for $CBOW$.
\end{itemize}
The source code pipeline has been inspired from «Chapter 11: The Word Embedding Model» from \cite{brownlee}.\raggedbottom

Figure \ref{fig1}, shows the visualization of Fon Word Embedding Space. We can see that words like \textit{t{\begin{tfour}\m{o}\end{tfour}} (father)} clusters close to \textit{fofó (big brother)}. We can also see the word \textit{fofó (big brother)} clusters close to \textit{sunnu (boy, man)}, while the word \textit{y{\begin{tfour}\m{o}\end{tfour}}nnúvi (girl, little girl)} clusters close to \textit{n{\begin{tfour}\m{o}\end{tfour}} (mother)}. We used the Fon Word2Vec to get the most similar words given positve references (see Table \ref{references_entities}).
\begin{figure}[t]
\includegraphics[width=\linewidth, height=70mm]{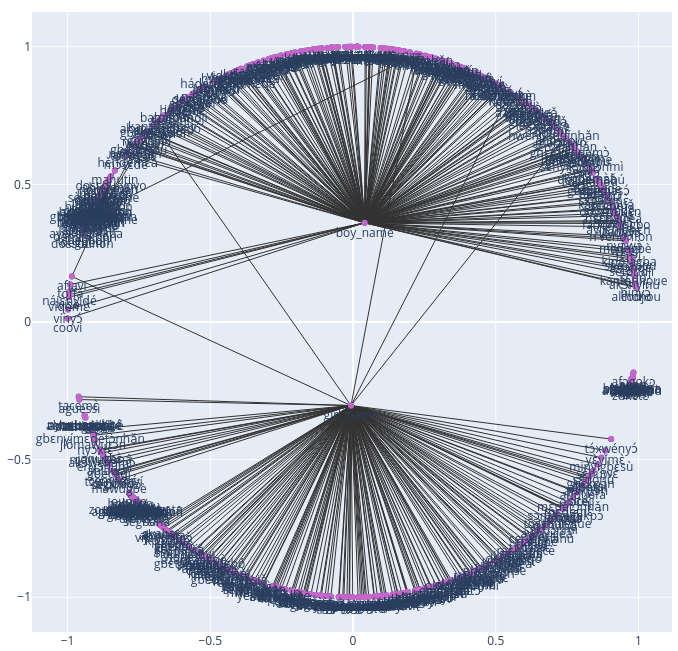}
\centering
\caption{\label{fig3} Fon Poincare Hierarchy Graph with constant negative curvature c = 10}
\end{figure}
\begin{figure}[t]
\includegraphics[width=\linewidth, height=70mm]{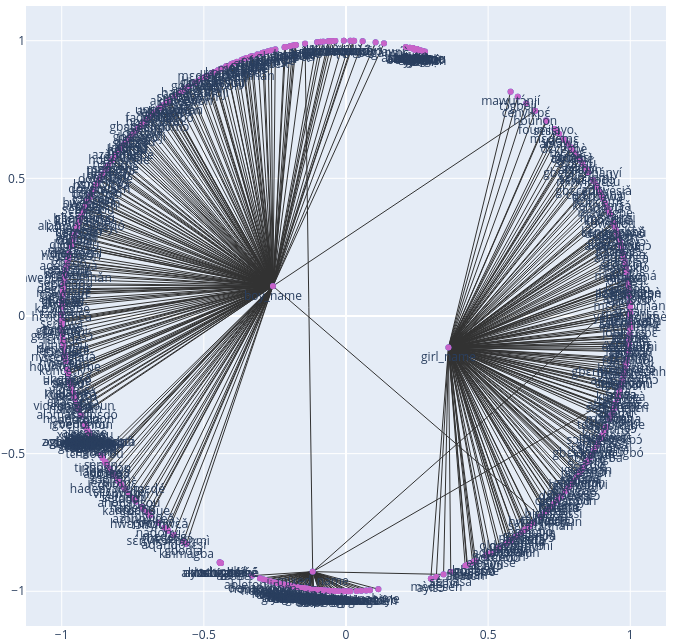}
\centering
\caption{\label{fig4} Fon Poincare Hierarchy Graph with constant negative curvature c = 15}
\end{figure}

One limitation of the Word2Vec is its restriction to the input corpora vocabulary. This makes the model, in case of very small corpora, very sensitive to out of vocabulary words, as similar representations can not be found or derived. 

However, these examples show that the model is able to correlate words among themselves taking into account their contexts. This proves the concept and importance of building WEs models for ALs words and entities, that could make easier NLP tasks on them.

\subsubsection{Word2Vec for Nobiin}
\label{nobinword2vec}
For Nobiin, we focus on the context of daily life style of family members, this to some extend reflects the culture in the geographic areas of this language. Our dataset contains nearly 40 sentences, with 7-50 words per sentence, reflecting the daily lifestyle contexts of Nobians.
We use word2Vec with the following hyperparameters:
\begin{itemize}[noitemsep]
   \item size: the number of dimensions of the embedding, set to $200$.
   \item \text{min\char`_count}: the minimum count of words to consider when training the model, set to $1$.
   \item $\alpha$: the learning rate set to $0.025$.
   \item window: the maximum distance between a target word and words around the target word, set to $15$.
   \item workers: the number of threads to use while training, set to $3$.
   \item sg: the training algorithm; $0$ for $CBOW$.
\end{itemize}
Figure \ref{fig2}, shows how the model represents relationships between different words along with their contexts. As we can see, the words (\textit{abô}, \textit{anên}, \textit{annenga}, \textit{annessi}) which are respectively defined as ($father$, $mother$, $brother$, $sister$) are clustered close to one another with the words ($Hassan$, $Umar$, $Nabra$, \textit{Munâ}) which are also personal names. We see also that words like (\textit{shây}, \textit{fatûr}) translated respectively as ($tea$, $breakfast$), very close to each other, and words like (\textit{tâjir}, $girish$, \textit{dukkân}) standing for ($merchant$, $money$, $supermarket$) clusters are close as well. The same remark is also applicable to other words like (\textit{semâ}, $masha$, $unatti$, $winji$) meaning respectively ($sky$, $sun$, $moon$, $stars$).

\begin{table}[t]
\begin{center}
\resizebox{\columnwidth}{!}{%
    \centering
\begin{tabular}{|p{4cm}|p{5cm}|p{2cm}|}
 \hline
 \textbf{References} & \textbf{Most similar representation} & \textbf{Similarity score}\\
 \hline
 n\begin{tfour}\m{o}\end{tfour}ví (sister, brother) &dadá (big sister)& 0.7928\\
\hline
n\begin{tfour}\m{o}\end{tfour}ví + sunnu (brother) &asi (wife)& 0.6626\\
 \hline
 t\begin{tfour}\m{o}\end{tfour} + ce (my father) & n\begin{tfour}\m{o}\end{tfour} (mother)& 0.6901\\
 \hline
  n\begin{tfour}\m{o}\end{tfour} (mother) & gle (crop field) & 0.6933\\
   \hline
  ny\begin{tfour}\m{o}\end{tfour}nuvi (little girl) & kp\begin{tfour}\m{e}\end{tfour}vi (small, little) & 0.8746\\
   \hline
  n\begin{tfour}\m{o}\end{tfour} + t\begin{tfour}\m{o}\end{tfour} (mother + father) & kpl\begin{tfour}\m{o}\end{tfour}n (education) & 0.7558\\
  \hline
\end{tabular}%
    }
\caption{\label{references_entities} Examples of most similiar representations in the Fon word2vec embedding space, given as input, positive references.}
\end{center}
\end{table}

\section{Fon and Nobiin Poincaré Embedding Models}
\label{poincare_section}
We also used $genism$ \footnote{https://radimrehurek.com/gensim/models/poincare.html} to implement Poincaré Embedding. 
To evaluate the model, we used the following criteria: the $mean$\text{\char`_}$rank$ $(MR)$, and the $Mean$ $Average$ $Precision$ $(MAP)$. The evaluation is done at two levels: 
\begin{itemize}[noitemsep]
\item $reconstruction$ which is defined as the capability of the observed data to reconstruct from the embedding. \citet{pointcare} defined it as a measure to evaluate the model's representation capacity.
\item $link$ $prediction$ to test generalization performance. 
\end{itemize}
Also the following modifications have been made to the source code of the model\footnote{https://github.com/alex-tifrea/poincare\text{\char`_}glove/blob/master/gensim/models/poincare.py}:
\begin{itemize}[noitemsep]
    \item the parameter $encoding$ = "$utf$-$8$" has been added as parameter to the csv reader object, used for the Link Prediction and Reconstruction file reading, to help them handle the diacritics (non-ascii characters).
    \item the function $find$\text{\char`_}$matching$\text{\char`_}$terms()$ to bring best possibilities in the model's vocabulary close to the input word, instead of considering only vocabulary words starting by the input word (as in the original code source), leading to $KeyError$ error.
\end{itemize}
Pull requests have been made to the official repository, and a repository containing the new version has been created too. The updated Poincaré model is available at: \url{https://github.com/bonaventuredossou/poincare_glove/blob/patch-3/gensim/models/poincare.py}.
\subsection{Poincaré Embedding Model for Fon}
\label{poincare_fon}
We implement Fon Poincaré Embedding on a dataset of names (boys, and girls or mixed names), benin cities, body parts, and date components (months of the year, days of the week). $Mixed$ names stand for names that could be attributed to both boys and girls. The dataset consists of a single unique relation among two different entities on each line following the HyperLex \citep{hyperlex} format, and contains 642 data samples. The dataset has been splitted into train (572 samples), validation (25 samples) and test (45 samples). Along the training dataset, we have \textit{218 boy names, 192 girl names, 67 mixed names, 43 benin cities, 12 months of the year and 7 days of the week}. All information and data entry of each entity, has been scraped respectively from external websites, and from crowd-sourcing through Google Form Surveys. For a better, understandable visualization and interpretation, we trained first the Fon Poincaré models with embedding space size set to 2. We believe that the concept could still be applied to higher dimensions that would however offer less understanding and interpretability. We tried also different constant negative curvatures ($10$, $15$). All the models have been trained on 2000 epochs.
The figures \ref{fig3}, and \ref{fig4} show different embedding spaces depending of the value of the constant negative curvature. We can notice clearly three different types of entities: \text{$boy$\char`_$name$}, \text{$girl$\char`_$name$} and \text{$mixed$\char`_$name$}. The \text{$benin$\char`_$city$} and \text{$body$\char`_$part$} entities components are all clustered together and really closed: elements of each group are, based on the graphs practically not distinguishable.
Considering the figures \ref{fig3}, and \ref{fig4} we can see that there are normal connections between $boy$\text{\char`_}$name$ and $mixed$\text{\char`_}$name$, as well as between $girl$\text{\char`_}$name$ and $mixed$\text{\char`_}$name$. However, on each figure, there exist $incorrect$ or $not$ $normal$ connections between $boy$\text{\char`_}$name$ and $girl$\text{\char`_}$name$ because both are not supposed to tie together, unless in case of \textit{mixed\text{\char`_}names}. We conclude that the constant negative curvature does not impact how groups intersect together but instead, impacts on the distance between elements among and across groups.

\begin{table}[]
\resizebox{\columnwidth}{!}{
    \centering
    \begin{tabular}{lllll}
    \hline
    \toprule
        \textbf{Dimensionality} & 2& 5 & 10 & 15\\
    \hline
    \midrule
        Reconstruction (MR/MAP) & 2.37/0.44 & 2.00/0.50 & 2.10/0.51 & \textbf{1.99/0.52}\\
        Link Prediction (MR/MAP) & 2.42/0.43 & 2.05/0.50 & 2.04/0.50 & \textbf{2.00/0.51}\\
        \hline
    \bottomrule
    \end{tabular}
    }
    \caption{Mean Rank and Mean Average Precision for Reconstruction and Link Prediction for Fon}
    \label{results}    
\end{table}
\begin{table}[]
\resizebox{\columnwidth}{!}{
    \centering
    \begin{tabular}{llllll}
    \hline
    \toprule
        \textbf{Dimensionality} & 2& 5 & 10 & 15 & 20\\
    \hline
    \midrule
        Reconstruction (MR/MAP) & 3.03/0.35 & \textbf{2.00/0.50} & 2.00/0.50 & 2.00/0.50 & 2.0/0.5\\
        Link Prediction (MR/MAP) & 2.75/0.39 & \textbf{2.00/0.50} & 2.00/0.50 & 2.00/0.50 & 2.00/0.50\\
        \hline
    \bottomrule
    \end{tabular}
    }
    \caption{Mean Rank and Mean Average Precision for Reconstruction and Link Prediction for Nobiin}
    \label{results_3}    
\end{table}
We choose the best model among the two described above, which is the one with constant negative curvature \textit{c = 15}. We also tried many other dimension size (5, 10, 15) and evaluated them. The results of the evaluations can be seen in table \ref{results}.
From these results, we can conclude that the higher the dimension, the better are the model reproducibility and generalization capacities.
\begin{table}[]
\resizebox{\columnwidth}{!}{
    \centering
    \begin{tabular}{lllll}
    \toprule
        \textbf{} & \textbf{Precision} & \textbf{Recall} & \textbf{F1-Score} & \textbf{Support}\\
    \midrule
        $boy$\text{\char`_}$name$ & 57 & \textbf{63} & 60 & 27\\
        $girl$\text{\char`_}$name$ & \textbf{33} & 28 & 30 & 18\\
        \hline
        accuracy &  &  & \textbf{49} & 45\\
        macro avg & 45 & \textbf{45} & 45 & 45\\
        weighted avg & 47 & \textbf{49} & 48 & 45\\
    \hline
    \bottomrule
    \end{tabular}
    }
    \caption{Classification Report of the Predictions of Fon model on the Fon Testing Dataset}
    \label{results_1}
\end{table}
\begin{table}[]
\resizebox{\columnwidth}{!}{
    \centering
    \begin{tabular}{lllll}
    \toprule
        \textbf{} & \textbf{Precision} & \textbf{Recall} & \textbf{F1-Score} & \textbf{Support}\\
    \midrule
        $boy$\text{\char`_}$name$ & 50 & \textbf{86} & 63 & 7\\
        $girl$\text{\char`_}$name$ & 50 & 14 & 22 & 7\\
        \hline
        accuracy &  &  & \textbf{50} & 14\\
        macro avg & 50 & \textbf{50} & 43 & 14\\
        weighted avg & 50 & \textbf{50} & 43 & 14\\
    \hline
    \bottomrule
    \end{tabular}
    }
    \caption{Classification Report of the Predictions of Nobiin model on the  Nobiin Testing Dataset}
    \label{results_4}
\end{table}
\begin{table}[]
\resizebox{\columnwidth}{!}{
    \centering
    \begin{tabular}{lllll}
    \toprule
        \textbf{} & \textbf{Precision} & \textbf{Recall} & \textbf{F1-Score} & \textbf{Support}\\
    \midrule
        $boy$\text{\char`_}$name$ & 62 & \textbf{96} & 75 & 27\\
        $girl$\text{\char`_}$name$ & \textbf{67} & 11 & 19 & 18\\
        \hline
        accuracy &  &  & \textbf{62} & 45\\
        macro avg & \textbf{64} & 54 & 43 & 45\\
        weighted avg & \textbf{64} & 62 & 53 & 45\\
    \hline
    \bottomrule
    \end{tabular}
    }
    \caption{Classification Report of the Predictions of Nobiin model on the  Fon Testing Dataset}
    \label{results_6}
\end{table}

\begin{table}[]
\resizebox{\columnwidth}{!}{
    \centering
    \begin{tabular}{lllll}
    \toprule
        \textbf{} & \textbf{Precision} & \textbf{Recall} & \textbf{F1-Score} & \textbf{Support}\\
    \midrule
        $boy$\text{\char`_}$name$ & 51 & \textbf{91} & 66 & 23\\
        $girl$\text{\char`_}$name$ & \textbf{50} & 9 & 15 & 22\\
        \hline
        accuracy &  &  & \textbf{51} & 45\\
        macro avg & \textbf{51} & 50 & 41 & 45\\
        weighted avg & 51 & \textbf{51} & 41 & 45\\
        \hline
    \bottomrule
    \end{tabular}
    }
    \caption{Classification Report of the Predictions of Fon model on the Nobiin Testing Dataset}
    \label{results_2}
\end{table}
We continued the experiments on with Fon using as dimension size \textit{d = 15}. Next, we used the model to predict entities types using the $score$\text{\char`_}$function()$ function, that computes the predicted score; extent to which a word $a$ is of entity type $b$.
For the predictions, we chose to focus on the entities: $boy$\text{\char`_}$name$, $girl$\text{\char`_}$name$, $body$\text{\char`_}$part$, $benin$\text{\char`_}$city$.
The entities $body$\text{\char`_}$part$ and $benin$\text{\char`_}$city$ ended up being predicted only either as $boy$\text{\char`_}$name$ or $girl$\text{\char`_}$name$, but more as $boy$\text{\char`_}$name$ than $girl$\text{\char`_}$name$. Therefore, we decided to drop them and only consider names ($boy$\text{\char`_}$name$ and $girl$\text{\char`_}$name$) entities. The model achieved an accuracy of 49\%. A closer look at the classification report (see table \ref{results_1}), shows that the current model is better at predicting boy names than girl ones.

Collecting more data on various entities types, training on more epochs, with higher constant negative curvature, or on higher dimensions could improve the model's performance and prediction capacity. \citet{pointcare} also showed that Poincaré embeddings are very successful in the embedding of large taxonomies, with regard to their representation capacity and their generalization performance.
\subsection{Poincaré Embedding Model for Nobiin}
For Nobiin, we implement a Poincaré embedding model with different constant negative curvatures (10, 15, 20, 25, 30) on a small dataset of nobian names (boys and girls). The dataset is a single level relationship among entities in each line following the HyperLex format \citep{hyperlex}. It consists of 108 data samples (training samples: 84 and test samples: 24). All the models have been trained on 2000 epochs.

\begin{figure}[t]
\includegraphics[width=\linewidth, height=70mm]{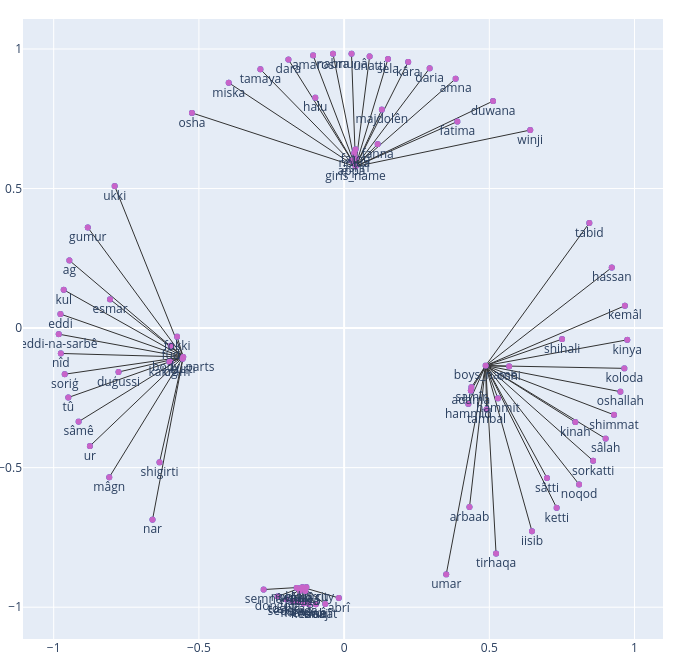}
\centering
\caption{\label{fig5} Nobiin Poincare Hierarchy Graph with constant negative curvature c = 15}
\end{figure}

\begin{figure}[t]
\includegraphics[width=\linewidth, height=70mm]{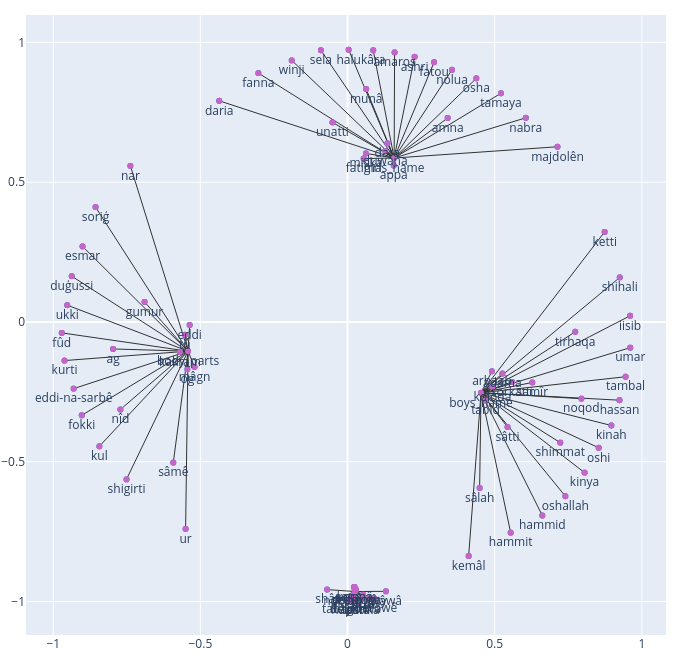}
\centering
\caption{\label{fig6} Nobiin Poincare Hierarchy Graph with constant negative curvature c = 20}
\end{figure}

We evaluated these models using the criteria described at the beginning of this section \ref{poincare_section}. We found that the best representations have been obtained for the models with constant negative curvature 15 and 20 (see figures \ref{fig5} and \ref{fig6}). To continue the experiments with Nobiin, we used the model with constant negative curvature \textit{c = 20}.

The Table \ref{results_3} shows the results of link prediction and reconstruction for the all trained models. The best results are already achieved from dimension size d = 5. For all further experiments with Nobiin, we choose as dimension size \textit{d = 10}.

The Table \ref{results_4} shows the classification report, with a global accuracy of 50\% of the best model built with constant negative curvature c = 20, and dimension size d = 10.

\subsection{Transfer Learning of Poincaré Embedding Models}
\label{joint}
Incentivized by the performance of transfer learning and its contribution to the state-of-the-art on a wide range of NLP tasks, we tested the Fon Poincaré Embedding model on 45 Nobian names, and the Nobiin Poincaré Embedding model on 45 Fon names.

In another words, we want to investigate how good, the information learned by the Fon and Nobiin Poincaré models, would be to improve the prediction of the Fon and Nobiin entities. To that extend, we evaluated the Nobiin Poincaré model on the Fon testing dataset: we got 62\% as global accuracy: the model predicted well most of $boy$\text{\char`_}$name$ and some of $girl$\text{\char`_}$name$ (see Table \ref{results_6}).

Next, we evaluated the Fon Poincaré model on the Nobiin testing dataset. As reported on Table \ref{results_2}, the Fon Poincaré model achieved an overall accuracy of 51.11\%. We can also see from that the Fon model predicted well the Nobian $boy$\text{\char`_}$name$ and some of $girl$\text{\char`_}$name$.

Despite the very small size of the datasets, one important remark here is the \textbf{\textit{improvement of the classification accuracy after transfer learning}}: the accuracy of classification of the Fon Poincaré model improved on the Nobiin test dataset, and likewise for the Nobiin Poincaré model on the Fon dataset. However, there is a need of building larger and contextualized corpora, to check on a more extended scale, the veracity of these promising results, for the African and low-resourced languages NLP research communities.

Thinking about reproducibility, transfer learning, and accompanied by will and hope, that the findings of this study could be extended to as many African Languages as possible, we can say that the information the models respectively and solely learned about the Fon and Nobiin data, helped them to predict averagely well $boy$\text{\char`_}$name$ and $girl$\text{\char`_}$name$ in the other language: both models used properties learned about the data they have been trained on, and extended it to the other language, to get the entities types right.

\section{An Approach to Linguistic and Social Interpretations}
\label{interpretation}
Unlike the $boy$\text{\char`_}$name$, we noticed a low $Recall$ of $girl$\text{\char`_}$name$ in the Poincaré models. As those results could infer some disparities, we tried to look for some linguistic, social interpretations. We found a plausible explanation could be that \textit{indigenous girl names} share a lot similarities with \textit{indigenous boy names} (mainly derivated from them), like the names \textit{ah{\begin{tfour}\m{o}\end{tfour}}ví (prince)} and \textit{ah{\begin{tfour}\m{o}\end{tfour}}ssí (princesse)}, sharing the same root.

\begin{figure}[t]
\includegraphics[width=\linewidth, height=70mm]{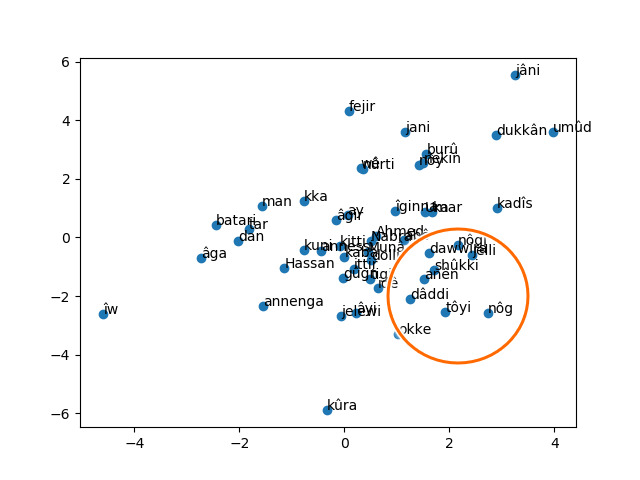}
\centering
\caption{\label{fig7}Visualization of Biased in one of Nobiin Word2Vec Embedding Models}
\end{figure}

WEs models can illustrate semantic and syntactic relationships between words, but they are not without flaws. The Figure \ref{fig7} shows one of our models clustering words of \textit{housekeeping activities}: \textit{(floor-sweeping: tôyi), (washing: shûkki), (utensils: dâddi), (cooking: okke)} close to the word \textit{mother (anên)}. We mitigated this bias by fine-tuning the maximum distance between the current and predicted words within a sentence, in the model of section \ref{nobinword2vec}, which clustered words $mother$ with \textit{father (abô), sister (annessi), brother (annenga)}, and \textit{housekeeping activities} words with the word \textit{house (nôg)}.

\citet{word2vec_bias}, also demonstrated in their study how WEs reinforced gender stereotypes at an alarming rate. They tested how Google would translate sentences from Turkish, which uses gender neutral pronouns, to English. Mostly, when a sentence contained descriptors stereotypically attributed to women \textit{(cook, teacher, nurse)}, the Turkish gender-neutral pronoun $o$ was translated to $she$. Conversely, sentences with terms such as \textit{hard working, lawyer}, and $engineer$ saw the pronoun translated as $he$.

\citet{word2vec_bias} and \citet{debias_multi} (inspired by \citep{word2vec_bias}), proposed efficient debiasing methods (hard and soft debiasing) respectively for \textit{binary class} (eg. male, female) and for \textit{multiclass} (gender, race, religion etc.) settings.

We want to precise that our point is not to blame Google but instead to emphasize that algorithms are based on a corpus of human words containing billions of data points. So WEs are merely reflecting already existing biases in societies. As our African open source WE models evolve, working on reducing disparities and inequalities would be a great future work pathway.
\section{Conclusion and Future Work}
In this paper we showed the possibilities and importance of building proper word embedding models for African Languages words and entities. We created Word2Vec and Poincaré Embedding models, for Fon and Nobiin, and showed that they can successfully represent similarities and relationships among those words and entities. We also presented promising transfer learning results, from the models accross the languages. The models we proposed, albeit on very small datasets, can further highly be improved, since some of the relationships have not been well generalized. Nevertheless, we believe that our findings are headrooms to developing better words embedding models, which could highly ease downstream NLP tasks and challenges on African Languages.
\section{Acknowledgments}
Authors would like to thank all contributors that helped gathering the data needed for the current study, especially Fabroni Bill Yoclunon, Ricardo Ahounvlame, and Nerry Koukoui.
\newpage
\bibliography{anthology, eacl2021}

\begin{thebibliography}{17}
\expandafter\ifx\csname natexlab\endcsname\relax\def\natexlab#1{#1}\fi

\bibitem[{Alabi et~al.(2019)Alabi, Amponsah-Kaakyire, Adelani, and
  España-Bonet}]{adelani}
Jesujoba~O. Alabi, Kwabena Amponsah-Kaakyire, David~I. Adelani, and Cristina
  España-Bonet. 2019.
\newblock \href {http://arxiv.org/abs/arXiv:1912.02481} {Massive vs. curated
  word embeddings for low-resourced languages. the case of yorùbá and twi}.

\bibitem[{Bolukbasi et~al.(2016)Bolukbasi, Chang, Zou, Saligrama, and
  Kalai}]{word2vec_bias}
Tolga Bolukbasi, Kai-Wei Chang, James Zou, Venkatesh Saligrama, and Adam Kalai.
  2016.
\newblock \href {http://arxiv.org/abs/arXiv:1607.06520} {Man is to computer
  programmer as woman is to homemaker? debiasing word embeddings}.

\bibitem[{Brownlee(2017)}]{brownlee}
Jason Brownlee. 2017.
\newblock \emph{Deep Learning for Natural Language Processing}.
\newblock Machine Learning Mastery.

\bibitem[{Devlin et~al.(2018)Devlin, Chang, Lee, and
  Toutanova}]{bert_embeddings}
Jacob Devlin, Ming-Wei Chang, Kenton Lee, and Kristina Toutanova. 2018.
\newblock \href {http://arxiv.org/abs/arXiv:1810.04805} {Bert: Pre-training of
  deep bidirectional transformers for language understanding}.

\bibitem[{Dossou and Emezue(2020)}]{ffr}
Bonaventure F.~P. Dossou and Chris~C. Emezue. 2020.
\newblock \href {http://arxiv.org/abs/arXiv:2006.09217} {Ffr v1.1: Fon-french
  neural machine translation}.

\bibitem[{Dossou et~al.(2021)Dossou, Yoclounon, Ahounvlamè, and
  Emezue}]{ffr_dataset}
Bonaventure F.~P. Dossou, Fabroni Yoclounon, Ricardo Ahounvlamè, and Chris
  Emezue. 2021.
\newblock \href {https://doi.org/10.5281/zenodo.4432712} {Fon french daily
  dialogues parallel data}.

\bibitem[{Manzini et~al.(2019)Manzini, Lim, Tsvetkov, and Black}]{debias_multi}
Thomas Manzini, Yao~Chong Lim, Yulia Tsvetkov, and Alan~W Black. 2019.
\newblock \href {http://arxiv.org/abs/arXiv:1904.04047} {Black is to criminal
  as caucasian is to police: Detecting and removing multiclass bias in word
  embeddings}.

\bibitem[{Mikolov et~al.(2013)Mikolov, Chen, Corrado, and Dean}]{word2vec}
Tomas Mikolov, Kai Chen, Greg Corrado, and Jeffrey Dean. 2013.
\newblock \href {http://arxiv.org/abs/arXiv:1301.3781} {Efficient estimation of
  word representations in vector space}.

\bibitem[{Nickel and Kiela(2017)}]{pointcare}
Maximilian Nickel and Douwe Kiela. 2017.
\newblock \href {http://arxiv.org/abs/arXiv:1705.08039} {Poincaré embeddings
  for learning hierarchical representations}.

\bibitem[{Norouzi et~al.(2014)Norouzi, Mikolov, Bengio, Singer, Shlens, Frome,
  Corrado, and Dean}]{norouzi2014zeroshot}
Mohammad Norouzi, Tomas Mikolov, Samy Bengio, Yoram Singer, Jonathon Shlens,
  Andrea Frome, Greg~S. Corrado, and Jeffrey Dean. 2014.
\newblock \href {http://arxiv.org/abs/1312.5650} {Zero-shot learning by convex
  combination of semantic embeddings}.

\bibitem[{Orife(2018)}]{orife}
Iroro Orife. 2018.
\newblock \href {http://arxiv.org/abs/arXiv:1804.00832} {Attentive
  sequence-to-sequence learning for diacritic restoration of yorùbá language
  text}.

\bibitem[{Pennington et~al.(2014)Pennington, Socher, and Manning}]{glove}
Jeffrey Pennington, Richard Socher, and Christopher Manning. 2014.
\newblock \href {https://doi.org/10.3115/v1/D14-1162} {{G}lo{V}e: Global
  vectors for word representation}.
\newblock In \emph{Proceedings of the 2014 Conference on Empirical Methods in
  Natural Language Processing ({EMNLP})}, pages 1532--1543, Doha, Qatar.
  Association for Computational Linguistics.

\bibitem[{Ravasz and Barab\'asi(2003)}]{barabasi}
Erzs\'ebet Ravasz and Albert-L\'aszl\'o Barab\'asi. 2003.
\newblock \href {https://doi.org/10.1103/PhysRevE.67.026112} {Hierarchical
  organization in complex networks}.
\newblock \emph{Phys. Rev. E}, 67:026112.

\bibitem[{Sappadla et~al.(2016)Sappadla, Nam, Loza~Menc{\'{\i}}a, and
  F{\"{u}}rnkranz}]{esann16zeroshot}
Prateek~Veeranna Sappadla, Jinseok Nam, Eneldo Loza~Menc{\'{\i}}a, and Johannes
  F{\"{u}}rnkranz. 2016.
\newblock \href
  {https://www.elen.ucl.ac.be/Proceedings/esann/esannpdf/es2016-174.pdf} {Using
  semantic similarity for multi-label zero-shot classification of text
  documents}.
\newblock In \emph{Proceedings of the 23rd European Symposium on Artificial
  Neural Networks, Computational Intelligence and Machine Learning (ESANN-16)},
  Bruges, Belgium. d-side publications.

\bibitem[{Vulić et~al.(2016)Vulić, Gerz, Kiela, Hill, and
  Korhonen}]{hyperlex}
Ivan Vulić, Daniela Gerz, Douwe Kiela, Felix Hill, and Anna Korhonen. 2016.
\newblock \href {http://arxiv.org/abs/arXiv:1608.02117} {Hyperlex: A
  large-scale evaluation of graded lexical entailment}.

\bibitem[{van~der Westhuizen and Niesler(2017)}]{isiZulu_wordEmbeddings}
Ewald van~der Westhuizen and Thomas Niesler. 2017.
\newblock \href {https://doi.org/10.21437/Interspeech.2017-1437} {Synthesising
  isizulu-english code-switch bigrams using word embeddings}.

\bibitem[{Xian et~al.(2017)Xian, Lampert, Schiele, and Akata}]{zeroshot}
Yongqin Xian, Christoph~H. Lampert, Bernt Schiele, and Zeynep Akata. 2017.
\newblock \href {http://arxiv.org/abs/arXiv:1707.00600} {Zero-shot learning --
  a comprehensive evaluation of the good, the bad and the ugly}.

\end{thebibliography}
\bibliographystyle{acl_natbib}

\end{document}